# Time Matters: Examine Temporal Effects on Biomedical Language Models

Weisi Liu, BS[1], Zhe He, PhD[2], Xiaolei Huang, PhD[1]
[1]University of Memphis, Memphis, TN, USA;
[2] Florida State University, Tallahassee, FL, USA

**Abstract**

*Time roots in applying language models for biomedical applications: models are trained on historical data and will be deployed for new or future data, which may vary from training data. While increasing biomedical tasks have employed state-of-the-art language models, there are very few studies have examined temporal effects on biomedical models when data usually shifts across development and deployment. This study fills the gap by statistically probing relations between language model performance and data shifts across three biomedical tasks. We deploy diverse metrics to evaluate model performance, distance methods to measure data drifts, and statistical methods to quantify temporal effects on biomedical language models. Our study shows that time matters for deploying biomedical language models, while the degree of performance degradation varies by biomedical tasks and statistical quantification approaches. We believe this study can establish a solid benchmark to evaluate and assess temporal effects on deploying biomedical language models.*

**Introduction**

Time naturally exists in the process of developing and deploying language models to diverse biomedical applications (e.g., disease diagnosis[1], biological reasoning[2], medical dialogues[3] and question answering[4]) that those biomedical models are usually trained on historical data while to be applied on future data or new patients. Biomedical language models (e.g., ClinicalBERT[5], BioRoBERTa[6], and ClinicalTransformer[7]), which were continuously trained language models on diverse biomedical text corpora, have achieved state-of-the-art performance on many domain tasks[8], including patient-outcome prediction[9], named entity recognition[10], and question answering[4]. However, time has rarely been considered in the model development and evaluation pipeline explicitly that the current pipeline randomly splits data into training, development, and test sets without considering data temporal shifts between development and deployment. Thus, a notable and common question in developing and deploying a biomedical model is: *How data temporal shifts will affect biomedical language model performance and generalizability, and if so, why*?

However, understanding the temporal effects on model performance is understudied on biomedical tasks, and the relation between data drifts and model performance is not clear. Unlike structured data (e.g., tabular data), unstructured data require extra steps to process and conversion to numerical feature representations, posing a concrete question of how to quantify unstructured data drifts. Common feature extraction approaches include three major directions, n-gram[11], word embeddings[12], and document embeddings[13], which convert unstructured texts into numerical vectors similar to structured data. While several existing studies (e.g., news[13, 11], political bills[11], and social media[13, 11]) have explored temporal effects on model performance, the studies primarily focus on general domains instead of biomedical domains, the focus of our study. Thus, a critical question is: *will patterns of the temporal drifts and effects in biomedical domain differ from the general domains, and if so, by how?* A recent study[14] presented observations of data drifts on biomedical classifiers. However, examining relations between data drifts and model performance expect diverse biomedical tasks and statistical evidences to better interpret observational findings, which have not been fully studied.

In this study, we aim to bridge the gaps by conducting well-designed experiments on state-of-the-art biomedical language models, systematically measuring data drifts, and statistically examining relations between data shifts and performance variations across three major biomedical tasks with domain-specific downstream tasks, phenotype inference (text classification), named entity recognition (information extraction), and question answering (text generation). We treat time spans as **domains** to segment data into subsets, train biomedical language models on historical domains, and evaluate models on future domains, such as training models by 2020 domain and evaluating models by 2024 domain. Our experiments include and deploy diverse state-of-the-art biomedical language models, such as ClinicalBERT[5], ClinicalT5[15], and BioRoBERTa[6], which are further fine-tuned on the three downstream tasks. We measure data drifts by diverse data encoding approaches and conduct statistical analysis with similarity and distance metrics to systematically examine temporal effects on model performance and relations between data shifts and performance variations. Our results have shown that data drifts do impact on model performance over time, yet, such effects may vary across biomedical tasks and biomedical language models. Our study establishes a new benchmark and a standard pipeline to understand and evaluate temporal effects on biomedical language models[1].

**Related Work**

**Model generalizability**, a critical yet unresolved challenge in the biomedical field, is to keep model perform well across various scenarios and settings. While increasing biomedical tasks have developed language models, the complex and heterogeneous nature of biomedical data pose a unique challenge for generalizing language models[16, 17]. For example, a recent study[18] showed that data variation can impact on model performance and in-domain or task-specific data can significantly boost performance of pre-trained biomedical language models by fine-tuning. Indeed, model generalizability is a broad concept in the biomedical field due to its data diversity and heterogeneity, such as generalizability for data imbalance[19] and demographic fairness[20, 21]. For data imbalance, Wu et al.[22] examines and shows that the token imbalance can cause underfitting on medical tokens and lower qualities of radiology reports; Shi et al.[23] explores the imbalance patterns of FDA drug datasets and finds overfitting issues of BERT-based classifiers[24] on majority labels. However, biomedical data usually spans over periods of **time** or longitudinal, and how the time factor impacts on the generalizability of biomedical language models has not been fully studied. Our study can fill the gaps by examining performance variations of language models over time and conducting analysis across critical biomedical tasks, including phenotype inference, named entity recognition, and question answering.

**Time** is implicitly embedded in building language models for biomedical tasks: biomedical language models are often built to be applied to future data that doesn't yet exist, and performance on held-out data is measured to estimate performance on future data whose distribution may have changed. A recent study[13] systematically examines temporal effects on model performance on the tasks of text classification and named entity recognition and suggests that model deterioration is task dependent and may not necessarily occur on pre-trained language models. However, those existing studies[12, 13, 25] examining temporal effects focus on the general domains (e.g., Wikipedia and news articles), and temporal effects on biomedical language models have rarely been studied. Thus, a similar yet underexplored question is that: *will temporal shifts of biomedical data show different patterns of impacting on the domain-specific language models, and if so, by how?* One line of close work[26, 27, 14] explored performance temporal variations of biomedical language models on the text classification task. However, it is unknown how data shifts impact on biomedical model performance beyond the classification task, which will be examined in this study with statistical evidence. Our work differs from and complements previous studies in the following aspects: 1) we examine temporal effects of Biomedical Language Models across multiple core biomedical tasks, 2) we explore the underlying reason of model temporal effect in the view of data temporal shifts, 3) we systematically quantify data shifts over time using diverse similarity and distance metrics, and 4) we analyze the relationship between measured data drifts and observed performance changes.

**Methods**
*Datasets and Tasks*

To systematically study the temporal effects on biomedical language models, we selected three time-varying datasets that meet three specific criteria: each data is a standard biomedical benchmark to evaluate model performance, each data should include time information (e.g., timestamps), the data should possess sufficient non-artificially generated data volumes, as detailed in Table 1. We specifically avoided LLM-annotated data to exclude the model collapse factor, where performance gradually deteriorates when models are trained on the data generated from other LLMs[28].

**MIMIC-IV-Note** (MIMIC) is a collection of de-identified clinical notes (discharge summary and radiology reports) as a part of the MIMIC-IV clinical database. The de-identification mechanism assigns the time information of each note to a three-year-long time interval spanning from 2008 to 2022, which naturally forms five temporal domains. We selected the first four temporal intervals for our experiments due to data scarcity in the last time interval. We choose the discharge summary and focus on the phenotype inference, a task to predict International Classification of Diseases (ICD) codes per discharge summary. The ICD code indicates the presence of diseases, symptoms, injuries, and other health conditions. Our phenotype inference task considers the top 50 frequent ICD-10 codes. For notes that only have ICD-9 labels, we convert the annotations into ICD-10 using the ICD-Mappings[30] toolkit.

**BioNER** is a biomedical information extraction task from the BioNLP Shared Task[31], which spans across different years. The corpora contain annotated biomedical entities primarily derived from PubMed abstracts. We selected 4 tracks with sufficient data entries and protein annotations: BioNLP09 Shared Task (2009ST)[31], BioNLP11 Epigenetics and Post-translational Modifications (2011EPI)[32], BioNLP11 Infectious Diseases (2011ID)[32], and BioNLP13 Genia Event Extraction (2013GE)[33]. We treat their release years as time domains, such as 2009 vs 2011 domains. Each BioNER data has five tags, IOBES (Inside, Outside, Beginning, End, Single), which delineate the boundaries of protein entities.

---
[1] Our code is available at https://github.com/trust-nlp/TemporalAssessment

**BioASQ** contains biomedical question answering data from 2013 to 2023[34]. Each question links to relevant snippets from PubMed articles and gold standard answers crafted by domain experts. We divided the data into four temporal domains by year and set the task as generating answers from the context snippets. Each snippet serves as a context per question, while exact answers were used as references for evaluating biomedical model performance.

**Table 1.** Overview of three time-varying biomedical data with four columns: intervals, task, labels, and data size.

| Dataset | Time Intervals | Task | Labels | Data Size |
|---------|---------------|------|--------|-----------|
| MIMIC | 2014-2016, 2008-2010, 2011-2013, 2017-2019. | Phenotype Inference | Top 50 frequent ICD codes | 331,794 notes |
| BioNER | 2009ST, 2011EPI, 2011ID, 2013GE. | Information Extraction | IOBES Tags of Protein Entity | 49,354 entities |
| BioASQ | 2013-2015, 2016-2018, 2019-2020, 2021-2023. | Question Answering | Gold Standard Answer | 5,046 questions |

*Language Models and Performance Evaluation*

**Biomedical Language Models:** For each task and data, we evaluated model performance and estimated temporal effects by diverse state-of-the-art language models. For the classification tasks of ICD-10 prediction on MIMIC-IV-Note and protein entity extraction on BioNER, we included the original BERT model[24] and models pre-trained on biomedical text corpora, including ClinicalBERT[5], BlueBERT[35], and BioRoBERTa[6]. For the generative question answering task on BioASQ, we included the T5 model[36] and the T5-based clinical variant ClinicalT5[15]. Ultimately, for each task, we selected the best-performing model to examine temporal effect.

**In-Time-domain and Cross-Time-domain evaluations:** We split each data into four time-intervals and treat each interval as a time domain, namely T1, T2, T3, and T4, shown in Table 1. To mitigate the impact of training data volume on performance, we sampled down the data volume for each temporal domain to match the size of the smallest domain. Additionally, we held out 20% of the data in each domain as a test set and used the rest for training and hyperparameter tuning. To assess temporal effect, we trained the model on a specific temporal domain and tested its performance on both the same temporal domain (*In-domain test*) and different temporal domains (*Cross-domain test*). We define the **temporal effect** as existing when there is a significant *performance change*, measured by the difference between cross-domain and in-domain test performances. For the cross-domain performance $p_{ij}$, we train a model on the training set of the time domain $T_i$ (source) and evaluate the test set of the time domain $T_j$ (target). For the in-domain performance $p_{ii}$, we train and test a model on the training and test sets from the same time domain $T_i$. To measure temporal effects on the domain pair $T_i - T_j$, we subtracted the performance score $p_{ij}$ from the in-domain evaluation ($p_{jj}$) $T_j - T_j$. We repeated each experiment five times per dataset to determine the statistical significance of the temporal effects. If a temporal effect exists, we expect that the change is statistically significant.

*Data Shift Metrics—Word Level*

Biomedical corpora is likely evolving over time—changes in biomedical terminology, treatment and disease manifestations; new biological entities and events been discovered every year; outdated terms occurring less), resulting in *data shifts on the word level*. The data shift can potentially impact the performance of the biomedical language models trained on data from a particular time period when applied to new data. To quantify and analyze temporal effects of the word-level data shifts, we employ two measures: Jaccard similarity and TF-IDF cosine similarity.

**Jaccard similarity** measures the similarity between finite sets by the ratio of the intersection and the union of two sets $S_1$ and $S_2$, as shown in Eq. (1). The similarity measurement allows us to assess the data drift by word usage change between the token sets of every two temporal domains.

$$Jaccard(S_1, S_2) = \frac{|S_1 \cap S_2|}{|S_1 \cup S_2|} \quad (1)$$

**TF-IDF cosine similarity**[37] measures the cosine of the angle between two TF-IDF vectors, which are TF-IDF scores for a collection of tokens. The TF-IDF is the product of two statistics, term frequency and inverse document frequency, as shown in Eq. (2). Term frequency is the relative frequency of a term within a document. The inverse document frequency is the logarithmic scaled inverse fraction of the documents containing the term and measures how much information the term provides. We compute the average TF-IDF vectors of each temporal domain and calculate the cosine similarity between two domain TF-IDF vectors $V_1$ and $V_2$ as shown in Eq. (3).

$$\text{tf}(t,d) = \frac{f_{t,d}}{\sum_{t' \in d} f_{t',d}} \quad \text{idf}(t,D) = \log\frac{|D|}{|\{d \in D: t \in d\}|} \quad (2)$$

$$TF-IDF\ Cosine\ Similarity(V_1, V_2) = \frac{V_1 \cdot V_2}{\|V_1\|\|V_2\|} = \frac{\sum_{i=1}^{n} tfidf_{1_i} tfidf_{2_i}}{\sqrt{\sum_{i=1}^{n} tfidf_{1i}^2} \sqrt{\sum_{i=1}^{n} tfidf_{2i}^2}} \quad (3)$$

*Data Shift Metrics—Semantic Level*

Semantic drift is another aspect of data drift, which can affect robust understanding of questions[38] and documents[12]. We aim to diversify our measurements of data drifts by estimating the semantic drifts over time. To estimate the semantic level of data shift, we employed diverse pre-trained neural models to extract semantic representations of each time domain, including the Universal Sentence Encoder (USE)[39], Sentence-BERT (SBERT)[40], BioLORD[41] and MedCPT[42].

**USE** translates sentences into high-dimensional vectors, encapsulating their semantic essence. Utilizing a transformer-based architecture, the encoder produces fixed-length representations of documents, effectively capturing their inherent semantic content. **SBERT** is a modification of the pre-trained BERT model[24] that use siamese and triplet network structures to derive semantically meaningful embeddings. The model expects that semantically similar documents are also close in the vector space. **BioLORD** is fine-tuned on sentence-transformers models using definitions and knowledge graph from biomedical domain. We chose their best-performing model, BioLORD-2023, for obtaining semantic embedding. **MedCPT** is a contrastive pre-trained transformer model trained with a large-scaled PubMed search logs. MedCPT retriever contains a query encoder and an article encoder that are trained to generate representations for search queries and PubMed articles, respectively. The model can effectively produce semantic embeddings for the BioNER and BioASQ datasets, which were from PubMed articles.

We obtain the embeddings for each temporal domain using the pre-trained models by calculating the average text embedding within that domain, creating a "*domain-average embedding*" that encapsulates the overall semantic distribution. To quantify data shifts across domains, we adopt the same metrics as used in the SBERT study[40], including cosine similarity, Euclidean distance, and Manhattan distance to measure semantic variations between domain-average embeddings.

*Statistical Verification*

We use diverse metrics to assess data shift between temporal domains, however, whether the data shift is significant and the relationship between data shift measurement and performance change are unclear. Therefore, we employ T-test to verify the statistical significance of performance change and data shift measurement, and we use Pearson correlation coefficient to quantify the relationship between data shift measurement and performance change.

**T-Test** is a statistical technique used to test whether the difference between the means of two groups is statistically significant or not. We apply the two-tailed T-test to test the statistical significance of performance change and data shift between each temporal domain pairs, with the null hypothesis that there is no significant performance change/

data shift between the source domain $T_i$ and target domain $T_j$. For *performance change*, we compare the cross-time-domain evaluations $\boldsymbol{p_{ij}}$ with the in-time-domain evaluations $\boldsymbol{p_{jj}}$, where $\boldsymbol{p_{ij}}$ and $\boldsymbol{p_{jj}}$ are the five-time observations in the experiments. If the p value is less than 0.05, the difference between the means of $\boldsymbol{p_{ij}}$ and $\boldsymbol{p_{jj}}$ is statistically significant, which means the performance change is significant. For *data shift* measurements, we randomly sample half of the data for each time domain and measure the data shift for all domain pairs $T_i - T_j$ to form cross-time-domain observations. And for in-time-domain pairs $T_j - T_j$, we randomly sample half of the data of the target domain $T_j$, and measure the similarity/ distance of the two halves as an in-time-domain observation. We generate 15 observations for each domain pairs and compare the cross-time-domain observations with the in-time-domain observations. If the p value is less than 0.05, we reject the null hypothesis at 95% confidence level, i.e. data shift measurement between the source domain $T_i$ and target domain $T_j$ is significant.

**Analysis**

In this section, we conduct statistical analyses to examine temporal effects on biomedical language models when data shift over time. While a recent study shows temporal effects on model performance are not statistically significant in the general domain (e.g., news and Wikipedia articles[13], very few studies have analyzed how data shifts may effect language model performance in the biomedical field, which *our analysis will fill the gap*. Our assumption is that if there is no statistically significant data shift, biomedical language models can perform consistently well over time. To verify our assumption, we treat *time intervals* as **domains** (e.g., 2014-2016 vs 2017-2019) and evaluate temporal effects by the performance change ($p_{jj} - p_{ij}$) by subtracting the in-domain test performance from the cross-domain test performance. The $p_{jj}$ refers to in-domain evaluation, and $p_{ij}$ refers to cross-domain evaluation, where the indices $i$ and $j$ are two time domains. Our analysis aims to answer three critical yet underexplored questions:

1. Does the performance of biomedical language models change over time?
2. Does the performance change statistically correlate with the data shift over time?
3. Do different shift measurements of biomedical data tell the same story?

*Q1: Does the performance of biomedical language models change over time?*

The dynamic nature of biomedical data, characterized by continuous evolution and temporal shifts, poses challenges in sustaining model efficacy over time. These shifts can result from various factors, including changes in disease prevalence, medical practices, and advancements in biomedical knowledge, potentially leading to significant variations in model performance. A recent research[14] evaluated temporal effects on classification model performance. However, the work focuses classification tasks and structured data, and the other biomedical tasks remain unexamined. We fill this gap by examining model performance temporal variation under three key biomedical tasks, including phenotype classification on MIMIC-IV, information extraction on BioNER, and question answering on BioASQ. To examine temporal effects, we investigated the model performance changes across time domain pairs and tasks.

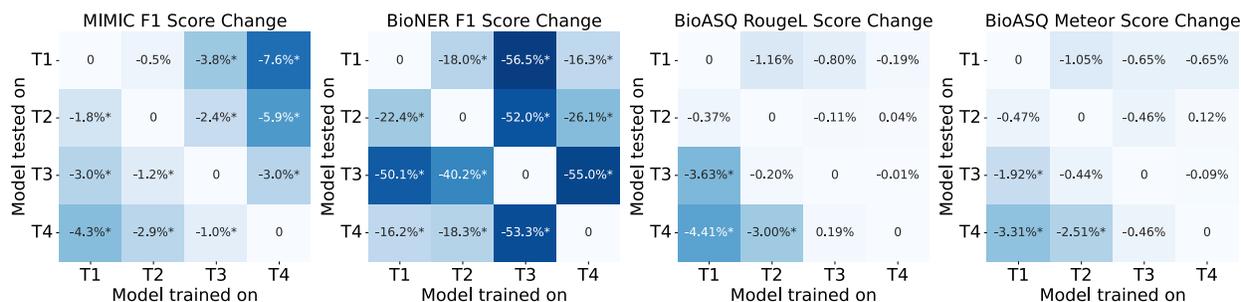

**Figure 1.** Performance changes as temporal effects. The star (*) indicates the performance change is significant.

We observed significant performance degradation in almost all cross-domain tests for MIMIC (phenotype inference) and BioNER (information extraction). But for BioASQ (question answering) when training models on historical data and performing test on future data, such as the performance degradation in the T1-T3, T1-T4, and T2-T4 evaluations in Figure 1. Notably, a larger time interval usually leads to a more performance decrease--more temporal effects happen with longer time intervals. While the reverse tests (T3-T1, T4-T1, and T2-T4) share similar patterns of temporal effects on MIMIC and BioNER, BioASQ did not show statistically significant performance degradation. We infer the asymmetry as for specific temporal events, the COVID-19. For example, the emergence of COVID-19 during the T3 period introduced questions and answers related to the pandemic in datasets T3 and T4. Models trained on earlier datasets lack the information to address these new queries effectively, whereas those trained on later data can still respond to earlier, more general questions. To verify this, we extracted the COVID-19 related data from the T4 domain test splits and evaluate the performances of models trained across different time domains. The RougeL scores for models trained from T1 to T4 are 0.376, 0.352, 0.388, and 0.419, respectively, and the Meteor scores for models trained from T1 to T4 are 0.204, 0.193, 0.210, and 0.259. These results confirm that *a specific temporal event can impact model performance across time*.

### Q2: Does the performance change statistically correlate with the data shift?

While both data and model performance shifts belong to temporal effects, it is unknown that if the two shifts correlate, implicating what can lead to the model performance drift. To answer the question, this section looks into the patterns of data shifts and model performance changes across temporal domain pairs in Figure 2. We can observe that any two domains with a closer temporal distance experience lower degree of data shifts and less performance declines. For example, in MIMIC dataset (first row of Figure 2), temporally adjacent domain pairs (e.g., T1-T2, T2-T3, and T3-T4) have higher similarity than non-adjacent pairs (e.g., T1-T3, T2-T4, and T1-T4). And in the cross-domain test on those adjacent domain pairs, model experienced relatively less performance declines (the first sub-figure of Figure 1). However, *temporal distance was not the determining factor for data shift and performance degradation.* Specifically, the major data shifts and performance declines are not limited to the domain pairs with the greatest temporal separation (T1-T4). For example, the most severe data shift happens on T1-T3 in BioNER dataset (the second row of Figure 2), along with the most drastic performance decline is observed in cross-domain test T3-T1, where model is trained on T3 and test performance on T1 data (the second sub-figure of Figure 2). *Notably, we observed a close correlation between data shifts and performance declines*. For example, the most severe data shift and performance degradation consistently occur within the same domain pair (e.g., T4-T1 for MIMIC, T3-T1 for BioNER).

We delve into a more precise statistical analysis of the correlation between performance changes and data shifts measured across domains by similarity and distance metrics. The results in Table 2 show that data shifts and performance changes are **statistically correlated** in the MIMIC and BioNER, but less correlated in the BioASQ, suggesting *temporal effects can vary across biomedical tasks*. We also observed that, for the same task, the correlation between different data shift measurements and performance changes can vary, as detailed in Table 2. This leads us to the next question: *Do All Data Shift Measurements Tell Us the Same Story?*

**Table 2.** The Pearson correlation coefficients between the performance changes and the data shift measurements. The star sign (*) indicates p-value is less than 0.05, and the double star sign (**) indicates p-value is less than 0.001.

| Metrics | MIMIC | | | BioNER | | | BioASQ | |
|---|---|---|---|---|---|---|---|---|
| | F1 | Precision | Recall | F1 | Precision | Recall | RougeL | Meteor |
| Jaccard Similarity | .68* | .71* | .58* | .77** | .65* | .74* | .36 | .45 |
| TF-IDF-Cosine | .74* | .79** | .62* | .94** | .74** | .89** | .51* | .56* |
| USE-Cosine | .70* | .76** | .59* | .91** | .72** | .86** | .53* | .53* |
| USE-Euclidean | -.74* | -.79** | -.62* | -.82** | -.67** | -.77** | -.54* | -.57* |
| USE-Manhattan | -.74* | -.79** | -.62* | -.82** | -.67** | -.77** | -.55* | -.57* |
| SBERT-Cosine | .86** | .90** | .75** | .95** | .74* | .90** | .54* | .54* |
| SBERT-Euclidean | -.87** | -.90** | -.75** | -.91** | -.73* | -.86** | -.52* | -.55* |
| SBERT-Manhattan | -.86** | -.90** | -.75** | -.91** | -.73* | -.86** | -.52* | -.55* |
| BioLORD-Cosine | .81** | .86** | .70* | .90** | .71* | .86** | .54* | .54* |
| BioLORD--Euclidean | -.83** | -.87** | -.71* | -.85** | -.69* | -.81** | -.52* | -.55* |
| BioLORD--Manhattan | -.83** | -.87** | -.71* | -.86** | -.69* | -.81** | -.52* | -.55* |
| MedCPT-Cosine | .86** | .90** | .75** | .97** | .75** | .92** | .62* | .62* |
| MedCPT-Euclidean | -.87** | -.91** | -.75** | -.95** | -.75** | -.90** | -.60* | -.62* |
| MedCPT-Manhattan | -.87** | -.91** | -.76** | -.95** | -.75** | -.90** | -.60* | -.62* |

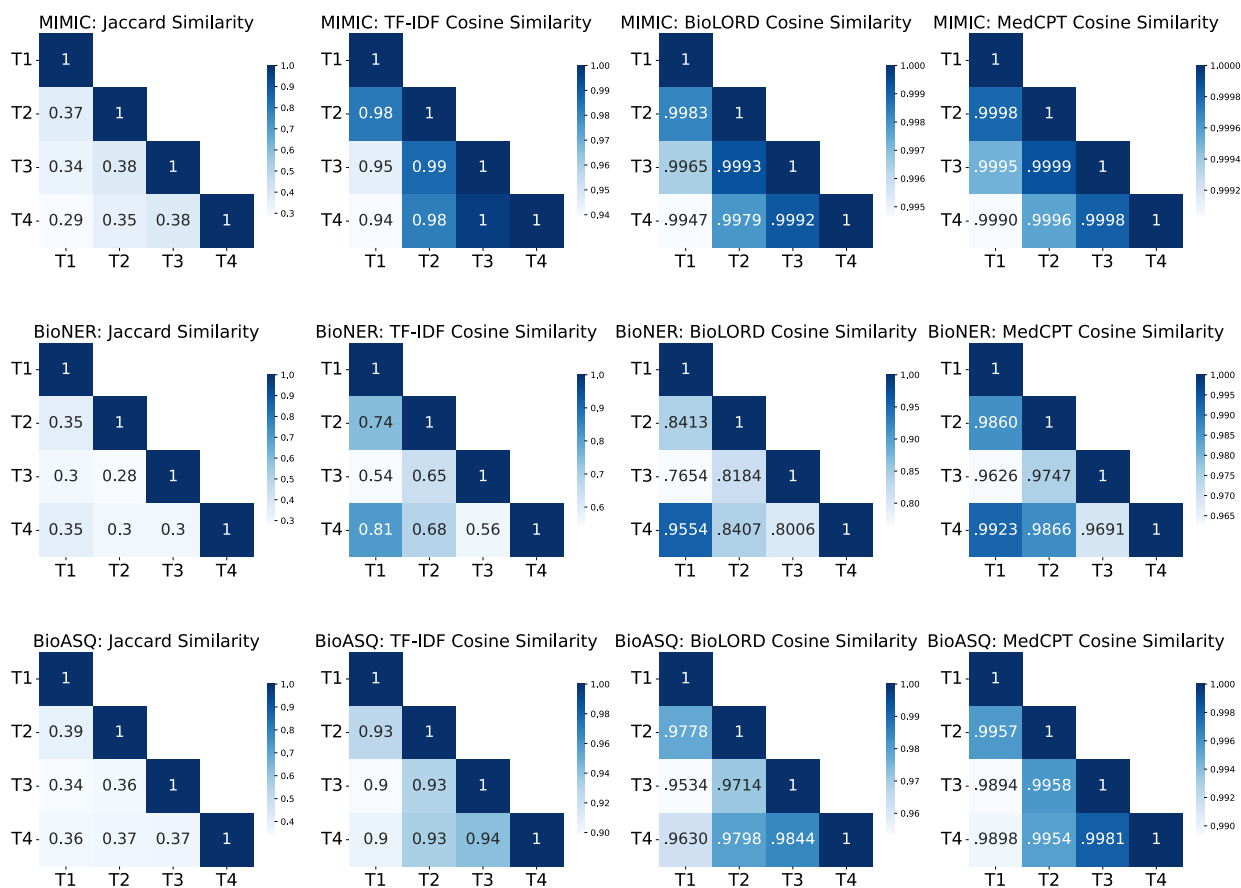

**Figure 2.** Heatmaps of data shift by similarity metrics only between time domains across MIMIC (first row), BioNER (second row), and BioASQ (last row) dataset. Lighter blue indicates lower similarity and more data variations.

*Q3: Do All Data Shift Measurements Tell Us the Same Story?*

Overall, the similarity metrics measuring word distribution shifts (first two rows in Table 2.) exhibit a lower correlation with performance changes than the distance metrics derived from domain-average embeddings, with lower correlation coefficients and higher p-values as shown in Table 2. However, these similarity metrics showed a stronger correlation for the BioNER dataset than MIMIC and BioASQ. The observation may suggest two implications: 1) the semantic level metrics (e.g., SBERT cosine similarity) are more effective than the word level ones (e.g., Jaccard similarity) at capturing the impact of data shifts on performance degradation; 2) while word level metrics remain effective for token-level tasks such as NER, they are less suitable for tasks requiring a lengthy context understanding, such as phenotype inference and question answering on long clinical notes.

Furthermore, among all distance metrics, those derived from domain-average embeddings and encoded by the MedCPT model generally showed higher correlation coefficients with performance degradation than metrics from general domain encoders like Universal Sentence Encoder[39]. Additionally, when measuring distances or similarity between the same domain-average embeddings, different metrics also exhibited variability in their correlation strengths. Cosine similarity receives the most consistently correlation.

**Conclusion**

Our study examined temporal effects on biomedical language models across various tasks, systematically evaluated how data drift affects model performance in biomedical applications. We developed a statistical framework to assess model performance variations over time and systematically measured data drift across different time periods. Our findings reveal that *while data drift does impact model performance, the extent of this influence varies depending on*

*the task*. Specifically, generative question answering tasks show resilience to data shifts, whereas information extraction tasks are more susceptible.

Furthermore, our analysis demonstrates that *different data shifts metrics provide varying perspectives on data drift, influencing model performance differently across tasks*. Word level similarity metrics, based on word distribution, are effective for classification tasks like named entity recognition but fall short in tasks requiring deeper understanding, such as document classification and question answering. In contrast, semantic level metrics based on domain-average embeddings show strong correlations across all tasks, with encoders trained on biomedical data offering more relevant insights than those trained on general domains.

**Acknowledgement**
The authors thank anonymous reviewers for their insightful feedback. This project was partially supported by the University of Memphis and College of Arts and Sciences and University of Florida Clinical and Translational Science Institute, which is supported in part by the National Science Foundation (NSF) under the award CNS-2318210 and IIS-2245920, National Institutes of Health (NIH) National Center for Advancing Translational Sciences under award UL1TR001427, as well as the Agency for Healthcare Research and Quality (AHRQ) under award R21HS029969. The content is solely the responsibility of the authors and does not necessarily represent the official views of the NSF, NIH, and AHRQ.

**References**
1. Kuroiwa T, Sarcon A, Ibara T, Yamada E, Yamamoto A, Tsukamoto K, et al. The potential of ChatGPT as a self-diagnostic tool in common orthopedic diseases: exploratory study. Journal of Medical Internet Research. 2023;25:e47621
2. Hsu CY, Cox K, Xu J, Tan Z, Zhai T, Hu M, et al. Thought graph: generating thought process for biological reasoning. In: Companion Proceedings of the ACM on Web Conference 2024; 2024. p. 537-40.
3. Chen Y, Zhao J, Wen Z, Li Z, Xiao Y. TemporalMed: advancing medical dialogues with time-aware responses in large language models. In: Proceedings of the 17th ACM International Conference on Web Search and Data Mining; 2024. p. 116-24.
4. Li D, Yang S, Tan Z, Baik JY, Yun S, Lee J, et al. DALK: dynamic co-ugmentation of LLMs and KG to answer Alzheimer's disease questions with scientific literature. arXiv preprint arXiv:240504819. 2024.
5. Alsentzer E, Murphy J, Boag W, Weng WH, Jindi D, Naumann T, et al. Publicly available clinical BERT embeddings. In: Rumshisky A, Roberts K, Bethard S, Naumann T, editors. Proceedings of the 2nd Clinical Natural Language Processing Workshop. Minneapolis, Minnesota, USA: Association for Computational Linguistics; 2019. p. 72-8. Available from: https://aclanthology.org/W19-1909.
6. Lewis P, Ott M, Du J, Stoyanov V. Pretrained language models for biomedical and clinical tasks: understanding and extending the state-of-the-art. In: Proceedings of the 3rd clinical natural language processing workshop; 2020. p. 146-57.
7. Yang X, Bian J, Hogan WR, Wu Y. Clinical concept extraction using transformers. Journal of the American Medical Informatics Association. 2020 10;27(12):1935-42. Available from: https://doi.org/10.1093/jamia/ocaa189.
8. Nerella S, Bandyopadhyay S, Zhang J, Contreras M, Siegel S, Bumin A, et al. Transformers and large language models in healthcare: A review. Artificial Intelligence in Medicine. 2024:102900.
9. Lyu W, Dong X, Wong R, Zheng S, Abell-Hart K, Wang F, et al. A multimodal transformer: fusing clinical notes with structured EHR data for interpretable in-hospital mortality prediction. In: AMIA Annual Symposium Proceedings. vol. 2022. American Medical Informatics Association; 2022. p. 719.
10. Naseem U, Khushi M, Reddy V, Rajendran S, Razzak I, Kim J. BioALBERT: a simple and effective pre-trained language model for biomedical named entity recognition. In: 2021 International Joint Conference on Neural Networks (IJCNN); 2021. p. 1-7.
11. Huang X, Paul M. Examining temporality in document classification. In: Proceedings of the 56th Annual Meeting of the Association for Computational Linguistics (Volume 2: Short Papers); 2018. p. 694-9.
12. Huang X, Paul MJ. Neural temporality adaptation for document classification: diachronic word embeddings and domain adaptation models. In: Korhonen A, Traum D, M`arquez L, editors. Proceedings of the 57th Annual Meeting of the Association for Computational Linguistics. Florence, Italy: Association for Computational Linguistics; 2019. p. 4113-23. Available from: https://aclanthology.org/P19-1403.
13. Agarwal O, Nenkova A. Temporal effects on pre-trained models for language processing tasks. Transactions of the Association for Computational Linguistics. 2022;10:904-21. Available from: https://aclanthology.org/2022.tacl-1.53.
14. Zhou H, Chen Y, Lipton Z. Evaluating model performance in medical datasets over time. In: Conference on Health, Inference, and Learning. PMLR; 2023. p. 498-508.

15. Lehman E, Johnson A. Clinical-t5: large language models built using mimic clinical text. PhysioNet. 2023.
16. Wang B, Xie Q, Pei J, Chen Z, Tiwari P, Li Z, et al. Pre-trained language models in biomedical domain: a systematic survey. ACM Comput Surv. 2023 oct;56(3). Available from: https://doi.org/10.1145/3611651.
17. Han G, Tsao J, Huang X. Length-aware multi-kernel transformer for long document classification. In: Bollegala D, Shwartz V, editors. Proceedings of the 13th Joint Conference on Lexical and Computational Semantics (*SEM 2024). Mexico City, Mexico: Association for Computational Linguistics; 2024. p. 278-90. Available from: https://aclanthology.org/2024.starsem-1.22.
18. Lehman E, Hernandez E, Mahajan D, Wulff J, Smith MJ, Ziegler Z, et al. Do we still need clinical language models? In: Mortazavi BJ, Sarker T, Beam A, Ho JC, editors. Proceedings of the Conference on Health, Inference, and Learning. vol. 209 of Proceedings of Machine Learning Research. PMLR; 2023. p. 578-97. Available from: https://proceedings.mlr.press/v209/eric23a.html.
19. De Angeli K, Gao S, Danciu I, Durbin EB, Wu XC, Stroup A, et al. Class imbalance in out-of-distribution datasets: Improving the robustness of the TextCNN for the classification of rare cancer types. Journal of Biomedical Informatics. 2022;125:103957. Available from: https://www.sciencedirect.com/science/article/pii/S1532046421002860.
20. Xu J, Xiao Y, Wang WH, Ning Y, Shenkman EA, Bian J, et al. Algorithmic fairness in computational medicine. eBioMedicine. 2022 oct;84:104250. Available from: https://linkinghub.elsevier.com/retrieve/pii/S2352396422004327.
21. Zhang J, Bandyopadhyay S, Kimmet F, Wittmayer J, Khezeli K, Libon DJ, et al. Developing a fair and interpretable representation of the clock drawing test for mitigating low education and racial bias. Scientific Reports. 2024;14(1):17444.
22. Wu Y, Huang IC, Huang X. Token imbalance adaptation for radiology report generation. In: Mortazavi BJ, Sarker T, Beam A, Ho JC, editors. Proceedings of the Conference on Health, Inference, and Learning. vol. 209 of Proceedings of Machine Learning Research. PMLR; 2023. p. 72-85. Available from: https://proceedings.mlr.press/v209/wu23a.html.
23. Shi Y, ValizadehAslani T, Wang J, Ren P, Zhang Y, Hu M, et al. Improving imbalanced learning by pre-finetuning with data augmentation. In: Moniz N, Branco P, Torgo L, Japkowicz N, Wozniak M, Wang S, editors. Proceedings of the Fourth International Workshop on Learning with Imbalanced Domains: Theory and Applications. vol. 183 of Proceedings of Machine Learning Research. PMLR; 2022. p. 68-82. Available from: https://proceedings.mlr.press/v183/shi22a.html.
24. Devlin J, Chang M, Lee K, Toutanova K. BERT: Pre-training of deep bidirectional transformers for language understanding. CoRR. 2018;abs/1810.04805. Available from: http://arxiv.org/abs/1810.04805.
25. Rijhwani S, Preotiuc-Pietro D. Temporally-informed analysis of named entity recognition. In: Jurafsky D, Chai J, Schluter N, Tetreault J, editors. Proceedings of the 58th Annual Meeting of the Association for Computational Linguistics. Online: Association for Computational Linguistics; 2020. p. 7605-17. Available from: https://aclanthology.org/2020.acl-main.680.
26. Harrigian K, Aguirre C, Dredze M. Do models of mental health based on social media data generalize? In: Cohn T, He Y, Liu Y, editors. Findings of the Association for Computational Linguistics: EMNLP 2020. Online: Association for Computational Linguistics; 2020. p. 3774-88. Available from: https://aclanthology.org/2020.findings-emnlp.337.
27. Guo LL, Pfohl SR, Fries J, Johnson AE, Posada J, Aftandilian C, et al. Evaluation of domain generalization and adaptation on improving model robustness to temporal dataset shift in clinical medicine. Scientific reports. 2022;12(1):2726.
28. Tan Z, Li D, Wang S, Beigi A, Jiang B, Bhattacharjee A, et al. Large language models for data annotation: a survey. arXiv preprint arXiv:240213446. 2024. Available from: https://arxiv.org/abs/2402.13446.
29. Johnson AEW, Bulgarelli L, Shen L, Gayles A, Shammout A, Horng S, et al. MIMIC-IV, a freely accessible electronic health record dataset. Scientific Data. 2023;10(1):1. Available from: https://doi.org/10.1038/s41597-022-01899-x.
30. Gonçalves S, Ehrensperger G. GitHub, editor. ICD-Mappings. GitHub; 2023. Available from: https://github.com/snovaisg/ICD-Mappings.
31. Kim JD, Ohta T, Tsujii J. Corpus annotation for mining biomedical events from literature. BMC bioinformatics. 2008;9:1-25.
32. Pyysalo S, Ohta T, Rak R, Sullivan D, Mao C, Wang C, et al. Overview of the ID, EPI and REL tasks of BioNLP shared task 2011. In: BMC bioinformatics. vol. 13. Springer; 2012. p. 1-26.
33. Kim JD, Wang Y, Yasunori Y. The genia event extraction shared task, 2013 edition-overview. In: Proceedings of the BioNLP Shared Task 2013 Workshop; 2013. p. 8-15.

34. Tsatsaronis G, Balikas G, Malakasiotis P, Partalas I, Zschunke M, Alvers MR, et al. An overview of the BIOASQ large-scale biomedical semantic indexing and question answering competition. BMC bioinformatics. 2015;16(1):1-28.
35. Peng Y, Yan S, Lu Z. Transfer learning in biomedical natural language processing: an evaluation of BERT and ELMo on ten benchmarking datasets. In: Proceedings of the 2019 Workshop on Biomedical Natural Language Processing (BioNLP 2019); 2019. p. 58-65.
36. Raffel C, Shazeer N, Roberts A, Lee K, Narang S, Matena M, et al. Exploring the limits of transfer learning with a unified text-to-text transformer. Journal of Machine Learning Research. 2020;21(140):1-67. Available from: http://jmlr.org/papers/v21/20-074.html.
37. Tata S, Patel JM. Estimating the selectivity of tf-idf based cosine similarity predicates. ACM Sigmod Record. 2007;36(2):7-12.
38. Chen Y, Xiao Y, Li Z, Liu B. XMQAs: constructing complex-modified question-answering dataset for robust question understanding. IEEE Transactions on Knowledge and Data Engineering. 2023.
39. Cer D, Yang Y, Kong Sy, Hua N, Limtiaco N, John RS, et al. Universal sentence encoder. arXiv preprint arXiv:180311175. 2018.
40. Reimers N, Gurevych I. Sentence-BERT: Sentence embeddings using siamese BERT-networks. In: Proceedings of the 2019 Conference on Empirical Methods in Natural Language Processing and the 9th International Joint Conference on Natural Language Processing (EMNLP-IJCNLP); 2019. p. 3982-92.
41. Remy F, Demuynck K, Demeester T. BioLORD-2023: semantic textual representations fusing large language models and clinical knowledge graph insights. Journal of the American Medical Informatics Association. 2024 02:ocae029. Available from: https://doi.org/10.1093/jamia/ocae029.
42. Jin Q, Kim W, Chen Q, Comeau DC, Yeganova L, Wilbur WJ, et al. MedCPT: contrastive pre-trained transformers with large-scale PubMed search logs for zero-shot biomedical information retrieval. Bioinformatics. 2023;39(11):btad651.